\pdfoutput=1  
\documentclass[11pt]{article}
\usepackage[hyperref]{acl}
\usepackage{times}
\usepackage{latexsym}
\usepackage{booktabs}
\usepackage[T1]{fontenc}
\usepackage{multirow}
\usepackage{amsmath,amssymb}
\usepackage{graphicx}
\usepackage{xcolor}
\usepackage{float}
\usepackage{url}
\usepackage{tikz}
\usepackage{pgfplots}
\pgfplotsset{compat=1.18}
\usetikzlibrary{shapes.geometric, arrows.meta, positioning, calc, fit, backgrounds, matrix}

\newcommand{\novel}{$\star$}

\title{Detecting Experiential Intertextuality Across Migration Routes:\\Beyond Surface Similarity in French Narratives}
\author{
  \textbf{Sakayo Toadoum Sari\textsuperscript{1}},
  \textbf{Nelly Robin\textsuperscript{2}},
  \textbf{Michelle Auzanneau\textsuperscript{2}},
  \textbf{Lakhdar Sais\textsuperscript{1}},
\\
  \textbf{V\'eronique Petit\textsuperscript{2}},
  \textbf{Marie Veniard\textsuperscript{3}},
  \textbf{Said Jabbour\textsuperscript{1}},
  \textbf{Fabien Delorme\textsuperscript{1}}
\\
\\
  \textsuperscript{1}CRIL, CNRS -- Universit\'e d'Artois, France,
  \textsuperscript{2}CEPED, Universit\'e Paris Cit\'e, France,
\\
  \textsuperscript{3}EDA, Universit\'e Paris Cit\'e, France
\\
  \small \textsuperscript{1}\texttt{\{sakayo,sais,jabbour,delorme\}@cril.fr},
  \small \textsuperscript{2}\texttt{nelly.robin@ird.fr},
\\
  \small \textsuperscript{2}\texttt{michelle.auzanneau@u-paris.fr},
  \small \textsuperscript{2}\texttt{veronique.petit@u-paris.fr},
\\
  \small \textsuperscript{3}\texttt{marie.veniard@u-paris.fr}
}

\begin{document}
\maketitle

\begin{abstract}
Migrants traversing geographically distinct routes such as the Trans-Saharan and Balkan corridors often recount strikingly parallel lived experiences: police violence, smuggler exploitation, dangerous crossings, and family separation. We introduce the task of \emph{experiential intertextuality detection}: automatically identifying shared experiential echoes across migration narratives without requiring annotated training data. From 108 French migration narratives spanning both corridors, we automatically generate sentence pairs and score them using annotation-free methods: lexical baselines, sentence embeddings, POS-based structural features, a migration-specific theme lexicon, context-aware narrative features, and zero-shot LLM scoring with Qwen2.5-7B and Mistral-7B under three prompting strategies. We validate all methods against 816 expert-annotated intertextuality judgments (inter-annotator Krippendorff's $\alpha{=}0.27$). Our results reveal that all surface, structural, and embedding methods correlate only weakly with expert judgments ($r{\leq}0.30$); Qwen2.5-7B zero-shot achieves the best single-method correlation ($r{=}0.38$); few-shot examples degrade Qwen but dramatically improve Mistral; narrative position significantly predicts intertextuality, with departure-phase pairs showing the highest experiential echoes; and a supervised hybrid combining all 31~features achieves $r{=}0.45$, a 21\% improvement over the best individual method.
\end{abstract}

\section{Introduction}
\label{sec:intro}

Migration is one of the defining phenomena of our era, and the narratives produced by migrants during their journeys are an invaluable yet underexploited source of knowledge for the Humanities and Social Sciences (HSS). These narratives collected through interviews at transit points along migration routes capture lived experiences in the migrants' own words: the dangers faced, the resources mobilized, and the complex decision-making that shapes each journey \cite{robin2014migrations, bacon2022fabrique}. A recurring observation among HSS researchers is that migrants following completely different geographical routes often describe strikingly similar experiences \cite{robin2014migrations, bacon2022fabrique}. A minor from C\^ote d'Ivoire crossing the Sahara and a Congolese refugee traversing the Balkans may both recount police violence at borders, exploitation by smugglers, perilous crossings, and the anguish of family separation. This phenomenon where distinct narratives echo shared experiential content despite originating from different geographical and cultural contexts is what we term \emph{experiential intertextuality}, a concept we
introduce in this paper to distinguish from traditional literary
intertextuality. We recognize experiential parallelism at three levels:
(i)~\emph{thematic}, when both sentences describe the same category
of experience (smuggler exploitation, police violence, dangerous
crossing); (ii)~\emph{functional}, when both occupy the same role in
the narrative arc of the journey (departure motivation, transit
danger, arrival); and (iii)~\emph{pragmatic}, when both carry the
same speech act or stance (self-motivation to leave, testimony of
suffering, expression of hope). Two sentences may share all three
levels while sharing no vocabulary whatsoever.

\begin{figure*}[t]
\centering
\begin{tikzpicture}[
    node distance=5mm and 12mm,
    data/.style={
        rectangle, rounded corners, draw=black!60, fill=orange!10,
        minimum width=2.8cm, minimum height=0.85cm, align=center, font=\scriptsize
    },
    method/.style={
        rectangle, rounded corners, draw=black!60, fill=blue!8,
        minimum width=2.8cm, minimum height=0.85cm, align=center, font=\scriptsize
    },
    result/.style={
        rectangle, rounded corners, draw=black!60, fill=green!10,
        minimum width=2.8cm, minimum height=0.85cm, align=center, font=\scriptsize
    },
    valid/.style={
        rectangle, rounded corners, draw=black!60, fill=red!10,
        minimum width=2.8cm, minimum height=0.85cm, align=center, font=\scriptsize
    },
    arrow/.style={-Latex, thick},
    darrow/.style={-Latex, thick, dashed},
    coltitle/.style={font=\small\bfseries, text=black!70}
]
\node[coltitle] at (0, 0.7) {Data Preparation};
\node[coltitle] at (5.5, 0.7) {Scoring Methods};
\node[coltitle] at (11, 0.7) {Evaluation};

\node[data] (narr) at (0,0) {108 French Narratives\\(Balkan + Trans-Saharan)};
\node[data, below=of narr] (sent) {Sentence Extraction\\+ Tokenization};
\node[data, below=of sent] (pairs) {Pair Generation\\(cross + intra-route)};
\node[data, below=of pairs] (theme) {Auto Theme Labeling\\(15 categories)};
\node[data, below=of theme] (ctx) {Context Localization\\(position in narrative)};
\draw[arrow] (narr) -- (sent);
\draw[arrow] (sent) -- (pairs);
\draw[arrow] (pairs) -- (theme);
\draw[arrow] (theme) -- (ctx);

\node[method] (lex) at (5.5, 0) {Lexical Baselines\\(TF-IDF, Jaccard, BM25,\\theme-lexicon --- 7 scores)};
\node[method, below=of lex] (pos) {POS-Structural\\(n-gram, edit, dep.\ triple,\\verb frame --- 4 scores)};
\node[method, below=of pos] (ctxf) {Context-Aware \novel\\(position sim., phase match,\\theme density --- 5 scores)};
\node[method, below=of ctxf] (emb) {Sentence Embeddings\\(CamemBERT, MiniLM,\\LaBSE, e5 --- 4 scores)};
\node[method, below=of emb] (llm) {Zero-Shot LLMs\\(Qwen2.5 + Mistral,\\3 strategies --- 6 scores)};
\draw[arrow] (ctx.east) -- ++(0.6,0) |- (lex.west);
\draw[arrow] (ctx.east) -- ++(0.6,0) |- (pos.west);
\draw[arrow] (ctx.east) -- ++(0.6,0) |- (ctxf.west);
\draw[arrow] (ctx.east) -- ++(0.6,0) |- (emb.west);
\draw[arrow] (ctx.east) -- ++(0.6,0) |- (llm.west);

\node[result] (scores) at (11, -0.4) {31 Scores per Pair\\(annotation-free)};
\node[valid, below=8mm of scores] (expert) {816 Expert Judgments\\(validation only)};
\node[result, below=8mm of expert] (corr) {Correlation Analysis\\(Pearson $r$, Spearman $\rho$)};
\node[result, below=of corr] (hybrid) {Hybrid Ridge\\(supervised upper-bound,\\$r = 0.454$)};
\node[result, below=of hybrid] (analysis) {Journey-Phase\\+ Theme Analysis};
\draw[arrow] (lex.east) -- ++(0.6,0) |- (scores.west);
\draw[arrow] (llm.east) -- ++(0.6,0) |- (scores.west);
\draw[arrow] (scores) -- (expert);
\draw[arrow] (expert) -- (corr);
\draw[arrow] (corr) -- (hybrid);
\draw[arrow] (hybrid) -- (analysis);
\draw[darrow] (expert.east) -- ++(0.5,0) |- (hybrid.east);
\end{tikzpicture}
\caption{Pipeline overview. \textbf{Left}: sentence pairs are automatically generated from raw narratives with context localization. \textbf{Middle}: five annotation-free method families produce 31 scores per pair. \textbf{Right}: scores validated against expert judgments; supervised hybrid as upper-bound. \novel~= novel.}
\label{fig:pipeline}
\end{figure*}
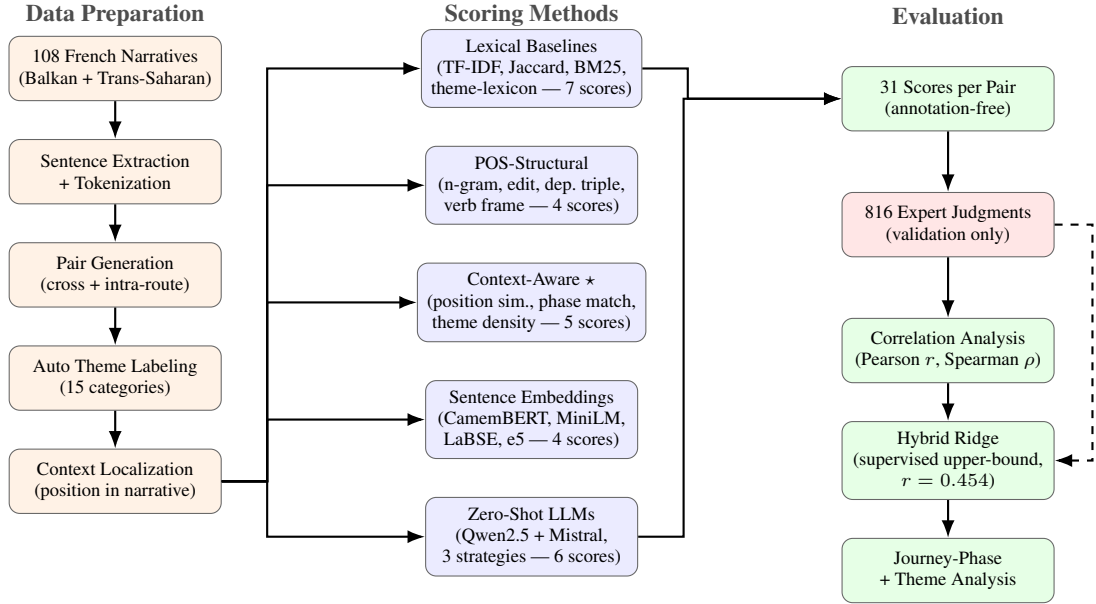

Unlike traditional intertextuality in literary studies, which concerns textual references and allusions between works \cite{kristeva1969semeiotike}, experiential intertextuality captures parallels in \emph{lived experience} as articulated through narrative. Detecting such parallels automatically is valuable for HSS researchers seeking to identify universal patterns in migration, understand which experiences transcend specific routes, and ultimately support policy-making with evidence-based insights. Prior work on these narratives \cite{ing2025textmining} focused on extracting domain terms and recognizing locations answering \emph{what} is mentioned. We address the complementary and harder question: \emph{do two sentences from different routes describe the same kind of experience?} This requires moving beyond entity extraction to experiential comparison, and from supervised evaluation against term lists to continuous correlation against expert judgments. We formalize this as a scoring task. Given a sentence pair $(s_i, s_j)$ drawn from narratives on different routes, we seek a function $f(s_i, s_j) \to [0, 1]$ that approximates expert-assessed experiential intertextuality, without requiring annotated training data. Figure~\ref{fig:pipeline} illustrates our approach.

Our contributions are: we formalize experiential intertextuality detection as a new NLP task with an annotation-free pipeline; we validate against 816 expert judgments with formal IAA; we introduce context-aware narrative features revealing journey-phase effects. Our code is publicly available.\footnote{\url{https://github.com/Toadoum/IntertextMigra}}

\section{Related Work}
\label{sec:related}

Text mining has been increasingly applied to migration-related texts, though primarily on public discourse. \citet{ozturk2018sentiment} use sentiment analysis on Twitter data to investigate public opinion toward the Syrian refugee crisis, while \citet{hussain2018analyzing} study shifts in blogosphere narratives during the European migrant crisis using named-entity extraction and targeted sentiment analysis. Both analyze \emph{public reactions} to migration, whereas we analyze \emph{migrants' own narratives}. Most closely related, \citet{ing2025textmining} present a text mining framework for migration narrative corpus focusing on domain term extraction via a modified set expansion algorithm (MultiWidthExpan) and location recognition/disambiguation using NER and BELA \cite{plekhanov2023multilingual}. Their evaluation uses P/R/F1 against expert term lists with no embedding or LLM baselines. Our work addresses a fundamentally different question on the same corpus: rather than extracting what entities are mentioned, we detect whether two sentences describe the same \emph{kind of experience}, evaluated via continuous correlation with expert intertextuality scores across 16 methods.

Semantic Textual Similarity (STS) is a well-established NLP benchmark \cite{cer2017semeval, agirre2016semeval}. Modern approaches leverage sentence embeddings from pretrained transformers \cite{reimers2019sentencebert, conneau2020unsupervised}, achieving strong performance on English STS benchmarks. However, experiential intertextuality is fundamentally different from semantic similarity: two sentences can be semantically dissimilar yet experientially parallel. For instance, \emph{On a pris le pickup pour aller à Gao} ("We took the pickup to go to Gao") and \emph{On a pris le bus pour aller à Belgrade} ("We took the bus to go to Belgrade") share the \emph{experience} of collective transit to a city, yet a standard STS model would score them low due to different locations and vehicle types. Conversely, sentences sharing keywords like "police" may receive high STS scores while describing entirely different experiences (routine checkpoint vs.\ violent refoulement).

Discourse relation frameworks such as RST \cite{mann1988rhetorical} and PDTB \cite{prasad2008penn} identify structural and semantic relations between text segments. Our task differs in that we compare segments \emph{across} documents rather than within a single document, and our relations are experiential rather than rhetorical. Computational narrative analysis has explored story similarity through emotional arcs \cite{reagan2016emotional}, narrative event chains \cite{chambers2008unsupervised}, and commonsense story understanding \cite{mostafazadeh2016corpus}. \citet{bamman2013learning} learn narrative schemas from text, while \citet{caselli2017event} propose event-centric approaches. Cross-document event coreference \cite{cybulska2014using} is close in spirit, comparing "the same kind of event" across documents, though it targets event identity rather than experiential parallelism. In digital humanities, quantitative intertextuality detection employs text reuse and sequence alignment for literary echoes \cite{forstall2015modeling}; our work extends this paradigm from textual allusion to experiential resonance. Our work adds a new dimension: identifying experiential parallels across narratives from different speakers in different geographical contexts, where the "events" are real lived experiences.

Recent work demonstrates that LLMs can perform nuanced discourse tasks, including stance detection, argumentation mining, and pragmatic interpretation \cite{gilardi2023chatgpt}. \citet{qwen2024} show strong multilingual capabilities for Qwen2.5, while the Mistral family \cite{jiang2023mistral} demonstrates competitive performance with efficient architectures. We evaluate both model families in a novel annotation-free setting, revealing substantial differences in how they handle experiential comparison.

\section{Data}
\label{sec:data}

Our corpus consists of 108 French-language migration narratives collected through semi-structured interviews at transit points along two corridors:

\begin{itemize}
    \item \textbf{Trans-Saharan corridor} (99 narratives): Collected in Niger (Agadez, Arlit), Algeria (Adrar, Tamanrasset, Maghnia), Senegal (Dakar, Mbour, Ziguinchor), and Morocco (Oujda, Rabat) from sub-Saharan minors. Countries of origin: C\^ote d'Ivoire, Mali, DRC, Guinea, Senegal, Nigeria, Gambia, Burkina Faso.
    \item \textbf{Balkan corridor} (9 narratives): Collected along transit routes through Serbia, North Macedonia, and Bosnia from migrants originating from Congo-Brazzaville, Algeria, Guinea, C\^ote d'Ivoire, Somalia, Mali, and Senegal.
\end{itemize}

Table~\ref{tab:corpus} reports corpus statistics. The narratives average 876 words, with Trans-Saharan narratives being shorter (avg.\ 787 words) due to the younger age of minors, while Balkan narratives are substantially longer (avg.\ 1,672 words) reflecting more complex multi-country journeys. The corpus was collected between 2015 and 2022 by trained researchers and local associations within the ANR HYCI project \cite{robin2014migrations} and related fieldwork, and was first used for computational analysis by \citet{ing2025textmining}. Interviews are conducted in French or local languages and transcribed into French. A relationship of trust is established through time and the guarantee of anonymity. Due to the sensitive nature of this data, we do not release it publicly (\S\ref{sec:ethics}).

\begin{table}[t]
\centering
\small
\caption{Corpus statistics by corridor. Countries = countries of origin of migrants.}
\label{tab:corpus}
\begin{tabular}{@{}lrrr@{}}
\toprule
& \textbf{Balkan} & \textbf{Trans-Sah.} & \textbf{Total} \\
\midrule
Narratives & 9 & 99 & 108 \\
Avg.\ words & 1{,}672 & 787 & 876 \\
Countries & 8 & 8 & 13 \\
Unique sentences & 3{,}592 & 2{,}330 & 5{,}922 \\
\bottomrule
\end{tabular}
\end{table}

To illustrate the nature of the data, consider these two excerpts from different corridors describing parallel experiences of smuggler exploitation:

\begin{quote}
\small
\textit{Trans-Saharan}: "J'ai pay\'e 75.000 FCFA pour passer en Alg\'erie. J'ai emprunt\'e le pickup avec une dizaine de migrants. On \'etait tr\`es serr\'e dans le pickup."
\emph{(I paid 75,000 FCFA to cross into Algeria. I took the pickup with about ten migrants. We were very cramped in the pickup.)}\\[3pt]
\textit{Balkan}: "Il m'avait vendu un passeport congolais avec un visa de la Turquie \`a l'int\'erieur [\ldots] quand je suis arriv\'e en Turquie, ils m'ont dit 'Monsieur ce n'est pas bon', et ils m'ont refoul\'e."
\emph{(He had sold me a Congolese passport with a Turkish visa inside [\ldots] when I arrived in Turkey, they told me 'Sir, this is not valid', and they deported me.)}
\end{quote}

Both describe exploitation, financial and documentary, yet share almost no vocabulary, precisely the challenge our task addresses. From the narratives, we automatically extract sentences and generate pairs using stratified sampling across three configurations: cross-route (one Balkan, one Trans-Saharan), intra-Balkan, and intra-Trans-Saharan. Each pair is automatically assigned thematic labels via keyword matching against a lexicon of 15 expert-defined categories, including \emph{violence\_police}, \emph{passeur\_exploitation}, \emph{travers\'ee\_dangereuse}, \emph{solidarit\'e}, \emph{famille\_s\'eparation}, \emph{exploitation\_travail}, \emph{document\_fraude}, \emph{d\'etention\_camp}, \emph{motivation\_d\'epart}, \emph{mineur\_seul}, \emph{r\^eve\_football}, \emph{discrimination\_racisme}, \emph{mort\_danger\_vital}, \emph{attente\_stagnation}, and \emph{autre}. No manual annotation is involved in pair generation.

A stratified sample of 816 pairs was annotated by two HSS migration experts on a continuous scale from 0.0 (no experiential link) to 1.0 (nearly identical experiences). Among these, 283 received independent dual annotations, yielding moderate agreement (Table~\ref{tab:iaa}). For dual-annotated pairs, the consensus score is the arithmetic mean; for the remaining 533 single-annotated pairs, the single annotator's score is used directly. No per-annotator normalization was applied, as both annotators' means are similar (0.389 vs.\ 0.405). This agreement level ($\alpha = 0.273$) is comparable to other subjective discourse annotation tasks \cite{prasad2008penn} and reflects genuine disagreement about what constitutes "shared experience."

\begin{table}[h]
\centering
\small
\caption{Inter-annotator agreement (AnnotatorA $\times$ AnnotatorB). The noise ceiling estimates the maximum attainable $r$ given annotator disagreement.}
\label{tab:iaa}
\begin{tabular}{lr}
\toprule
\textbf{Metric} & \textbf{Value} \\
\midrule
Dual-annotated pairs & 283 \\
Krippendorff's $\alpha$ (interval) & 0.273 \\
Cohen's $\kappa$ (weighted, 3-bin) & 0.259 \\
Pearson $r$ & 0.281*** \\
Mean $|$diff$|$ & 0.306 \\
Spearman-Brown reliability ($\hat{\rho}$) & 0.438 \\
\textbf{Noise ceiling} ($r_{\max} = \sqrt{\hat{\rho}}$) & \textbf{0.662} \\
\bottomrule
\end{tabular}
\end{table}

To contextualize our results, we estimate the maximum correlation any method can achieve given annotator disagreement. Using the Spearman-Brown prophecy formula, the reliability of the averaged score from two annotators is $\hat{\rho} = 2r_{12} / (1 + r_{12}) = 0.438$, where $r_{12} = 0.281$ is the inter-annotator Pearson~$r$. The noise ceiling, the theoretical maximum $r$ between a perfect method and the consensus score, is $r_{\max} = \sqrt{\hat{\rho}} = 0.662$. Our best single method (Qwen2.5 zero-shot, $r = 0.375$) achieves 56.6\% of this ceiling, and the supervised hybrid ($r = 0.454$) achieves 68.6\%, indicating substantial room for improvement but also that a significant portion of the remaining gap is attributable to irreducible annotator noise.

\section{Methods}
\label{sec:methods}

Let $\mathcal{N} = \{n_1, \ldots, n_K\}$ denote a corpus of $K$ narratives, each consisting of sentences $n_k = (s_1^k, \ldots, s_{m_k}^k)$. Given a pair $(s_i, s_j)$ from different narratives, we define scoring functions $f: \mathcal{S} \times \mathcal{S} \to [0, 1]$ that estimate experiential intertextuality without supervision.

\subsection{Lexical Baselines}
We compute seven lexical scores. Let $W(s)$ denote the word set of sentence~$s$. Beyond standard measures (Jaccard similarity, ROUGE-1 F-score, BM25-approximate, character 3-gram overlap, and overlap coefficient) we employ TF-IDF cosine similarity fitted on the full 108-narrative corpus rather than sentence pairs alone, for better inverse document frequency estimation. We additionally define a domain-specific \textbf{theme lexicon score}. We distinguish between the 15~thematic \emph{labels} used for pair categorization and a subset of $L=8$ thematic \emph{word sets} used for computing the lexicon feature; the 8 sets correspond to experiential categories with sufficient distinctive vocabulary (\emph{violence}, \emph{passeur}, \emph{transport}, \emph{famille}, \emph{danger}, \emph{travail}, \emph{document}, \emph{h\'ebergement}), totaling 130~terms curated from the narratives. Categories like \emph{autre} and \emph{r\^eve\_football} were excluded from the lexicon as they lack stable keyword indicators. The score is:
\begin{equation}
f_{\text{thm}}(s_i, s_j) = \frac{1}{L} \sum_{l=1}^{L} \mathbf{1}[W_i \cap T_l \neq \emptyset] \cdot \mathbf{1}[W_j \cap T_l \neq \emptyset]
\label{eq:theme}
\end{equation}
where $W_i = W(s_i)$, counting the fraction of thematic categories activated in \emph{both} sentences. This captures topical co-occurrence at the experiential category level rather than the word level.

\subsection{POS-Structural Features}
We hypothesize that migrants describing parallel experiences may use similar grammatical structures even with entirely different vocabulary. Let $P(s) = (p_1, \ldots, p_n)$ denote the POS tag sequence of~$s$ (punctuation removed), obtained via spaCy's French model (\texttt{fr\_core\_news\_sm}). We compute:

\noindent\textbf{POS n-gram Jaccard.} Let $G(s)$ be the multiset of POS $n$-grams ($n \in [2,5]$):
\begin{equation}
f_{\text{png}}(s_i, s_j) = \frac{\sum_{g} \min(G_i[g],\, G_j[g])}{\sum_{g} \max(G_i[g],\, G_j[g])}
\label{eq:posng}
\end{equation}
where $G_i = G(s_i)$ and $G_j = G(s_j)$.

\noindent\textbf{POS edit similarity.} Let $P_i = P(s_i)$:
\begin{equation}
f_{\text{ped}}(s_i, s_j) = 1 - \frac{\text{Lev}(P_i,\, P_j)}{\max(|P_i|,\, |P_j|)}
\label{eq:posed}
\end{equation}

\noindent\textbf{Dependency triple overlap.} Let $D(s) = \{(r, p_h, p_c)\}$ be the set of dependency triples (relation, head POS, child POS). We compute Jaccard over $D(s_i)$ and $D(s_j)$.

\noindent\textbf{Verb-frame overlap.} Using the dependency parse, for each verb in a sentence we extract the set of dependency labels of its children (ignoring the verb's lemma). This captures action argument structures cross-vocabulary: "prendre [obj, obl]" matches whether the object is "pickup" or "bus".

\subsection{Context-Aware Narrative Features}
Since each sentence $s_i$ originates from a specific narrative $n_k$ at a known position, we exploit source context, a signal unavailable to sentence-level methods.

\noindent\textbf{Position similarity.} Let $\pi(s_i) \in [0, 1]$ be the normalized character offset of $s_i$ within its source narrative (0 = beginning, 1 = end):
\begin{equation}
f_{\text{pos}}(s_i, s_j) = 1 - |\pi(s_i) - \pi(s_j)|
\label{eq:possim}
\end{equation}

The intuition is that migration narratives follow a natural temporal arc (departure, transit, arrival), and sentences at similar positions describe experiences from similar journey phases.

\noindent\textbf{Journey-phase match.} Let $\phi: [0,1] \to \{1,2,3,4\}$ map positions to four phases (quartiles): departure (1), early transit (2), late transit (3), arrival (4):
\begin{equation}
f_{\text{ph}}(s_i, s_j) = \begin{cases}
1.0 & \text{if } \phi_i = \phi_j \\
0.5 & \text{if } |\phi_i - \phi_j| = 1 \\
0.0 & \text{otherwise}
\end{cases}
\label{eq:phase}
\end{equation}
where $\phi_i = \phi(\pi(s_i))$.

\noindent\textbf{Context theme density.} For each sentence, we extract a 500-character window from the source narrative centered on the sentence's position, and compute the fraction of theme-lexicon words. The similarity of densities between two sentences captures whether both originate from thematically rich narrative passages.

\subsection{Sentence Embeddings}
We evaluate four multilingual sentence embedding models, listed with HuggingFace identifiers for reproducibility: \texttt{sentence-camembert-large} (CamemBERT-STS) \cite{martin2020camembert}; \texttt{paraphrase-multilingual-MiniLM-L12\\-v2} \cite{reimers2019sentencebert}; \texttt{LaBSE}; and \texttt{multilingual-e5-large} \cite{conneau2020unsupervised}. We use multilingual models to enable future extension to English narratives from the same project. For each pair, we compute cosine similarity between L2-normalized embeddings.

\subsection{LLM Scoring}
We evaluate two open-source 7B-parameter LLMs, \texttt{Qwen2.5-7B-Instruct} \cite{qwen2024} and \texttt{Mistral-7B-Instruct-v0.3} \cite{jiang2023mistral}, both NF4-quantized via \texttt{bitsandbytes}, with temperature $= 0.1$ and \texttt{max\_new\_tokens} $= 256$, under three prompting strategies:

\noindent\textbf{Zero-shot.} The system prompt defines experiential intertextuality in French and provides the 0--1 rating scale. The model receives only the two sentences and returns a JSON score. No examples are provided.

\noindent\textbf{Few-shot.} Three expert-annotated pairs are prepended as demonstrations, selected to cover low ($\sim$0.1), medium ($\sim$0.4), and high ($\sim$0.8) intertextuality scores. This tests whether calibration examples help the model understand the scale.

\noindent\textbf{Chain-of-thought (CoT).} The prompt requests structured reasoning: (1) identify the theme of each sentence, (2) compare whether the experiences are parallel, (3) produce a score. This tests whether explicit reasoning improves scoring quality.

An example of the zero-shot prompt structure:

\begin{quote}
\small
\texttt{System:} \textit{Tu es un expert en \'etudes migratoires (You are an expert in migration studies). \'Evalue le degr\'e d'intertextualit\'e exp\'erientielle entre deux phrases (Assess the degree of experiential intertextuality between two sentences) [\ldots] \'Echelle: 0.0 \`a 1.0 (Scale: 0.0 to 1.0). R\'eponds avec un JSON (Respond with a JSON object): \{"score": <float>\}}\\[2pt]
\texttt{User:} \textit{Phrase 1 (Sentence 1): "[sentence 1]" \\Phrase 2 (Sentence 2): "[sentence 2]"}
\end{quote}

\subsection{Hybrid Model (Supervised Upper-Bound)}
We combine all $d = 31$ features into $\mathbf{x}_{ij} \in \mathbb{R}^d$ for each pair and train Ridge regression with narrative-level grouped 5-fold CV (GroupKFold; pairs are grouped by narrative-pair ID so that no narrative appears in both train and test folds, preventing data leakage from shared sentences; see Appendix~\ref{app:sampling} for the annotation sampling protocol):
\begin{align}
\hat{y}_{ij} &= \mathbf{w}^\top \mathbf{x}_{ij} + b \label{eq:ridge}\\
\mathcal{L} &= \textstyle\sum_{(i,j)} (y_{ij} - \hat{y}_{ij})^2 + \lambda \|\mathbf{w}\|^2 \nonumber
\end{align}
where $y_{ij}$ is the expert score. The 31 features comprise: 7~lexical, 4~POS, 5~context-aware, 4~embedding, 6~LLM scores, plus \texttt{pos\_combined} (weighted POS combination), raw narrative positions ($\pi(s_i)$, $\pi(s_j)$), and two metadata indicators (\texttt{same\_country}, \texttt{cross\_route}). All 16 annotation-free methods (lexical, POS, context, embedding, LLM) are computed without access to expert labels; TF-IDF weights are fitted on the full 108-narrative corpus (analogous to using a pretrained model), not on CV folds. Unlike other methods, the hybrid \emph{uses expert labels} and serves as an upper-bound.

\section{Experiments and Results}
\label{sec:results}

All annotation-free methods are validated by Pearson~$r$ and Spearman~$\rho$ against the expert sample ($n=816$). No method uses expert scores for training except the supervised hybrid. Experiments were conducted on a GPU machine equipped with single NVIDIA Quadro RTX~8000 (48\,GB VRAM) using a virtual environment with PyTorch, Transformers, and sentence-transformers. Table~\ref{tab:main} presents the complete results ranked by Pearson~$r$.

\begin{table}[h]
\centering
\small
\setlength{\tabcolsep}{3pt}
\caption{Methods vs.\ expert validation ($n{=}816$). 95\% CIs for Pearson $r$ via Fisher $z$-transform. \novel~novel contribution of this paper. $^\dagger$~uses expert labels (narrative-grouped CV). **~$p{<}0.01$, ***~$p{<}0.001$. Note: 815/816 expert pairs are cross-route.}
\label{tab:main}
\begin{tabular}{@{}lccc@{}}
\toprule
\textbf{Method} & \textbf{$r$} & \textbf{95\% CI} & \textbf{$\rho$} \\
\midrule
\multicolumn{4}{l}{\textit{LLM (zero-shot / few-shot / CoT)}} \\
\quad Qwen2.5 zero-shot & \textbf{.375}*** & [.314, .432] & .304*** \\
\quad Qwen2.5 CoT & .365*** & [.304, .423] & \textbf{.331}*** \\
\quad Mistral few-shot & .327*** & [.264, .387] & .272*** \\
\quad Qwen2.5 few-shot & .289*** & [.224, .351] & .239*** \\
\quad Mistral CoT & .243*** & [.177, .307] & .219*** \\
\quad Mistral zero-shot & .134*** & [.066, .201] & .124*** \\
\midrule
\multicolumn{4}{l}{\textit{Sentence Embeddings}} \\
\quad multi-MiniLM & .295*** & [.231, .356] & .254*** \\
\quad CamemBERT-STS & .284*** & [.220, .346] & .236*** \\
\quad LaBSE & .249*** & [.184, .312] & .204*** \\
\quad e5-large & .219*** & [.153, .283] & .193*** \\
\midrule
\multicolumn{4}{l}{\textit{Lexical}} \\
\quad Theme-lexicon & .287*** & [.223, .349] & .270*** \\
\quad TF-IDF (original) & .254*** & [.189, .317] & .286*** \\
\quad Char-3gram & .230*** & [.164, .294] & .184*** \\
\midrule
\multicolumn{4}{l}{\textit{Context-aware} \novel} \\
\quad Position sim. & .131*** & [.063, .198] & .119*** \\
\quad Phase match & .107** & [.039, .174] & .102** \\
\quad Theme density (avg) & .089* & [.020, .157] & .074* \\
\quad Theme density (sim) & .049 & [$-$.020, .118] & .035 \\
\quad Window overlap & .011 & [$-$.058, .080] & .015 \\
\midrule
\multicolumn{4}{l}{\textit{POS-structural}} \\
\quad POS n-gram & .119*** & [.051, .186] & .076 \\
\quad Dep.\ triple & .091** & [.023, .159] & .061 \\
\midrule
\multicolumn{4}{l}{\textit{Supervised upper-bound}$^\dagger$} \\
\quad Hybrid Ridge (31) & \textbf{.454}*** & [.398, .507] & \textbf{.383}*** \\
\midrule
\multicolumn{4}{l}{\textit{Noise ceiling}} \\
\quad $r_{\max}$ & .662 & --- & --- \\
\bottomrule
\end{tabular}
\end{table}

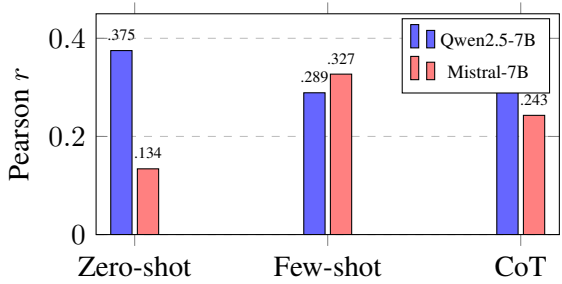
\begin{figure}[h]
\centering
\begin{tikzpicture}
\begin{axis}[
    ybar,
    bar width=8pt,
    width=\columnwidth,
    height=4.5cm,
    ylabel={Pearson $r$},
    symbolic x coords={Zero-shot, Few-shot, CoT},
    xtick=data,
    ymin=0, ymax=0.45,
    legend style={at={(0.98,0.98)}, anchor=north east, font=\scriptsize},
    ymajorgrids=true,
    grid style=dashed,
    every node near coord/.append style={font=\tiny},
    nodes near coords,
    nodes near coords align={vertical},
    point meta=explicit symbolic,
]
\addplot[fill=blue!60] coordinates {
    (Zero-shot, 0.375) [.375]
    (Few-shot, 0.289) [.289]
    (CoT, 0.365) [.365]
};
\addplot[fill=red!50] coordinates {
    (Zero-shot, 0.134) [.134]
    (Few-shot, 0.327) [.327]
    (CoT, 0.243) [.243]
};
\legend{Qwen2.5-7B, Mistral-7B}
\end{axis}
\end{tikzpicture}
\caption{Prompting strategy comparison. Few-shot degrades Qwen but dramatically improves Mistral.}
\label{fig:prompting}
\end{figure}

\textbf{No annotation-free method exceeds $r = 0.38$}, but contextualized against the noise ceiling of $r_{\max} = 0.662$ (\S\ref{sec:data}), Qwen2.5 achieves 56.6\% of the theoretical maximum.
The \textbf{theme lexicon} (Eq.~\ref{eq:theme}, $r = 0.287$) outperforms three of four neural embedding models; Qwen2.5 significantly outperforms it (Williams test: $t = 3.03$, $p = 0.003$).
\textbf{Qwen2.5-7B} strongly outperforms Mistral-7B in zero-shot ($0.375$ vs.\ $0.134$), a gap larger than any prompting strategy effect.
The \textbf{hybrid} ($r = 0.454$, 68.6\% of ceiling) significantly outperforms the best single method (Williams test: $t = 4.61$, $p < 0.0001$), demonstrating strong complementarity.
\textbf{POS features are weak} ($r \leq 0.119$), discussed in \S\ref{sec:discussion}.
Note: 815 of 816 expert pairs are cross-route, so results directly measure inter-corridor experiential echoes. Figure~\ref{fig:prompting} reveals a striking asymmetry between the two LLMs:

\begin{itemize}
    \item \textbf{Qwen2.5-7B}: Zero-shot is strongest ($r = 0.375$); few-shot \emph{degrades} performance to $r = 0.289$ ($-23\%$); CoT maintains strength ($r = 0.365$) and yields the best ranking quality ($\rho = 0.331$).
    \item \textbf{Mistral-7B}: Zero-shot is weakest ($r = 0.134$); few-shot \emph{dramatically improves} to $r = 0.327$; CoT is intermediate ($r = 0.243$).
\end{itemize}

This asymmetry suggests that Qwen has stronger zero-shot French comprehension and migration-relevant world knowledge, while Mistral requires calibration to understand the task. The few-shot degradation for Qwen may reflect \emph{score anchoring}: the three demonstration examples bias the model toward their score distribution, overriding its own superior zero-shot judgment. This finding has practical implications for annotation-free deployment: the choice of prompting strategy must be model-specific.

Figure~\ref{fig:heatmap} shows expert intertextuality stratified by journey phase. Departure $\times$ departure ($\mu = 0.280$) is highest, while mismatched phases (e.g., arrival $\times$ departure, $\mu = 0.129$) are lowest. The position similarity feature (Eq.~\ref{eq:possim}) captures this effect ($r = 0.131$, $p < 0.001$).

\begin{figure}[h]
\centering
\begin{tikzpicture}
\matrix[
    matrix of nodes,
    row sep=-\pgflinewidth,
    column sep=-\pgflinewidth,
    nodes={draw, minimum width=1.3cm, minimum height=1.0cm, anchor=center, font=\scriptsize},
    row 1/.style={nodes={fill=gray!15, font=\scriptsize\bfseries, minimum height=0.7cm}},
    column 1/.style={nodes={fill=gray!15, font=\scriptsize\bfseries, minimum width=1.4cm}},
] (m) {
              & Dep.   & E-Tr.  & L-Tr.  & Arr.   \\
|[fill=gray!15]| Dep.   & |[fill=red!35]| .280  & |[fill=red!18]| .227  & |[fill=red!15]| .218  & |[fill=red!10]| .194  \\
|[fill=gray!15]| E-Tr.  & |[fill=blue!8]| .129  & |[fill=red!25]| .263  & |[fill=red!17]| .224  & |[fill=red!10]| .189  \\
|[fill=gray!15]| L-Tr.  & |[fill=blue!8]| .128  & |[fill=blue!5]| .169  & |[fill=red!8]|  .175  & |[fill=red!20]| .243  \\
|[fill=gray!15]| Arr.   & |[fill=blue!8]| .129  & |[fill=blue!5]| .143  & |[fill=red!25]| .275  & |[fill=red!20]| .240  \\
};
\end{tikzpicture}
\caption{Mean expert intertextuality by journey-phase pair (row = sentence~1, column = sentence~2). Darker red = higher. The diagonal and near-diagonal show strongest echoes.}
\label{fig:heatmap}
\end{figure}

The diagonal pattern confirms that matching journey phases yield stronger intertextuality, with departure being the most universal. This aligns with HSS research observing that the motivations for leaving (family pressure, economic hardship, conflict) are shared across African migration contexts \cite{robin2014migrations}, while transit and arrival experiences are shaped by route-specific factors (desert vs.\ sea crossings, different border policies). Notably, the off-diagonal pair arrival $\times$ late\_transit ($\mu = 0.275$) also scores high, suggesting that experiences near the end of the journey converge regardless of exact phase boundaries: migrants describe similar exhaustion, hope, and encounters with authorities. Table~\ref{tab:themes} reports mean expert scores by thematic category. Life-threatening danger ($\mu = 0.340$) and labor exploitation ($\mu = 0.326$) show the strongest cross-route intertextuality, suggesting that these experiences are systemic to irregular migration regardless of corridor. Document fraud ($\mu = 0.286$) and the football dream ($\mu = 0.254$) also show strong echoes: forged papers and aspirations of an athletic career are remarkably consistent themes. In contrast, unaccompanied minor experiences ($\mu = 0.117$) and waiting/stagnation ($\mu = 0.125$) are the weakest, indicating route-specific variation. Waiting experiences depend heavily on local transit infrastructure (desert oases vs.\ Balkan refugee camps), while unaccompanied minor narratives reflect different legal frameworks across countries.

\begin{table}[H]
\centering
\small
\caption{Mean expert intertextuality by theme ($n \geq 30$).}
\label{tab:themes}
\begin{tabular}{@{}lrc@{}}
\toprule
\textbf{Theme} & \textbf{$n$} & \textbf{Expert $\mu$} \\
\midrule
mort\_danger\_vital & 30 & 0.340 \\
exploitation\_travail & 69 & 0.326 \\
document\_fraude & 57 & 0.286 \\
r\^eve\_football & 49 & 0.254 \\
travers\'ee\_dangereuse & 79 & 0.242 \\
passeur\_exploitation & 93 & 0.196 \\
violence\_police & 82 & 0.186 \\
d\'etention\_camp & 42 & 0.182 \\
solidarit\'e & 66 & 0.177 \\
attente\_stagnation & 123 & 0.125 \\
mineur\_seul & 46 & 0.117 \\
\bottomrule
\end{tabular}
\end{table}

Table~\ref{tab:ablation} decomposes the supervised hybrid (Eq.~\ref{eq:ridge}) into feature groups.

\begin{table}[h]
\centering
\small
\caption{Ablation (narrative-grouped 5-fold CV Ridge). LLM features contribute most ($\Delta r = -0.059$). Removing embeddings slightly \emph{improves} the model.}
\label{tab:ablation}
\begin{tabular}{@{}lccc@{}}
\toprule
\textbf{Configuration} & \textbf{$r$} & \textbf{$\rho$} & \textbf{$\Delta r$} \\
\midrule
Full (31 feat.) & .454 & .383 & --- \\
\midrule
\multicolumn{4}{l}{\textit{Each group alone:}} \\
\quad LLM (6) & .416 & .335 & --- \\
\quad Lexical (7) & .338 & .268 & --- \\
\quad Embedding (4) & .295 & .249 & --- \\
\quad POS (4) & .100 & .059 & --- \\
\midrule
\multicolumn{4}{l}{\textit{Leave-one-group-out:}} \\
\quad w/o LLM & .395 & --- & $-$.059 \\
\quad w/o Lexical & .430 & --- & $-$.024 \\
\quad w/o POS & .453 & --- & $-$.001 \\
\quad w/o Embedding & .463 & --- & +.010 \\
\bottomrule
\end{tabular}
\end{table}

Three findings emerge. First, \textbf{LLM features are the most valuable group}: removing them causes the largest drop ($\Delta r = -0.059$), and LLM features alone achieve $r = 0.416$, already higher than any non-LLM method. Second, \textbf{embeddings are redundant}: removing all four embedding models actually \emph{improves} the hybrid from $r = 0.454$ to $0.463$, indicating that the semantic similarity signal captured by embeddings is entirely subsumed by the LLM scores. Third, \textbf{POS features are negligible}: removal changes $r$ by only $0.001$, confirming that syntactic structure provides almost no independent signal for experiential intertextuality. The feature importance analysis (Ridge coefficients) shows that the top-5 most influential features are: ROUGE-1 ($\beta = -0.111$, a suppressor, since controlling for unigram overlap lets other features capture genuine experiential similarity), \emph{jaccard} ($\beta = +0.077$), \emph{theme\_lexicon} ($\beta = +0.045$), \emph{Qwen2.5-CoT} ($\beta = +0.041$), and \emph{ctx\_position\_sim} ($\beta = +0.036$).

To verify that LLMs capture signal beyond surface overlap, we compute partial correlations controlling for Jaccard similarity. The Qwen2.5-7B zero-shot correlation with expert scores drops only marginally when partialing out Jaccard ($r = 0.375 \to r_{\text{partial}} = 0.351$), confirming substantial beyond-surface signal. Similarly, partialing out the theme lexicon yields $r_{\text{partial}} = 0.336$, indicating that Qwen captures experiential parallels not reducible to thematic keyword overlap. For the theme lexicon itself, controlling for Jaccard yields $r_{\text{partial}} = 0.253$ (vs.\ raw $r = 0.287$), showing that approximately 12\% of its signal is explained by simple word overlap, with the remainder reflecting genuine thematic co-activation. We examine the pairs where all methods disagree most with experts to understand the limits of current approaches.

\paragraph{High expert, low predicted.} Experts rate these pairs as experientially parallel, but no method detects it. A representative example: "Mon p\`ere dit que je suis courageux et que je peux r\'eussir en Europe" (\emph{My father says I am brave and can succeed in Europe}; Trans-Saharan) and "Je me suis dit que je ne peux pas rester comme \c{c}a" (\emph{I told myself I cannot stay like this}; Balkan). Both express the moment of deciding to migrate, sharing no vocabulary, no syntactic structure, and no thematic keywords. This is the hardest case: experiential intertextuality encoded purely in \emph{pragmatic intent}, the speech act of self-motivation before departure.

\paragraph{Low expert, high predicted.} Methods score these pairs highly, but experts disagree. For example, sentences sharing "police" or "fronti\`ere" in different experiential contexts, such as a routine identity check versus a violent \emph{refoulement}. The same vocabulary describes fundamentally different experiences. This confirms that lexical overlap, and even embedding similarity, can mislead when identical words carry different experiential weight.

\paragraph{Annotator disagreement cases.} Among the 283 doubly-annotated pairs, those with the largest AnnotatorA--AnnotatorB disagreement ($|$diff$| > 0.6$) tend to involve implicit experiential links, such as one sentence describing a cause ("J'ai pay\'e le passeur") and another describing a consequence ("On \'etait 20 dans le pickup"), where recognizing the link requires domain knowledge about smuggling logistics.

\section{Discussion}
\label{sec:discussion}

\paragraph{Experiential intertextuality $\neq$ similarity.}
No standard similarity method achieves strong correlation with expert judgments. The best single method (Qwen2.5-7B, $r = 0.375$) explains only 14\% of variance ($r^2 = 0.141$). This is a property of the phenomenon: experiential parallels are expressed through different vocabulary, syntax, and discourse structure.

\paragraph{Domain knowledge $>$ neural embeddings.}
The theme lexicon ($r = 0.287$) outperforms CamemBERT-STS ($r = 0.284$), LaBSE ($r = 0.249$), and e5-large ($r = 0.219$). For domain-specific experiential analysis, 130 curated terms can rival neural models with millions of parameters.

\paragraph{LLMs: model identity matters more than prompting.}
The Qwen--Mistral gap in zero-shot ($0.375$ vs.\ $0.134$) is 2.8$\times$ larger than any within-model prompting effect. Pretraining data composition is more important than prompt engineering. The few-shot asymmetry shows that calibration examples serve different functions depending on model capability.

\paragraph{Journey phase as a structural predictor.}
The significant correlation of position similarity with expert scores ($r = 0.131$, $p < 0.001$) shows that migration narratives have an inherent temporal structure conditioning the universality of experiences \cite{robin2014migrations}.

\paragraph{Why POS features underperform.}
POS features yielded only weak correlations ($r \leq 0.119$) because POS patterns are too generic (e.g., \texttt{PRON VERB PREP NOUN} matches both relevant and irrelevant sentences) and varying French proficiency among migrants produces diverse syntax for identical experiences.

\paragraph{Toward event-structure representations.}
Our POS and dependency features capture only shallow syntax. Richer event-centric representations (semantic role labeling, frame-semantic parsing, or predicate-argument structures) could better capture the experiential content of sentences by abstracting over vocabulary while preserving "who did what to whom." For instance, an SRL-based feature could match the agent-action-patient structure of "Le passeur nous a abandonn\'es" and "The smuggler left us behind" despite lexical divergence. However, robust French SRL tools remain limited compared to English, and available frame-semantic resources (e.g., French FrameNet) offer incomplete coverage for migration-specific events such as \emph{refoulement}, smuggling, or border crossing. We consider event-structure baselines an important direction for future work, particularly as multilingual SRL models improve.

\paragraph{Implications for HSS research.}
The universality of life-threatening danger ($\mu = 0.340$) and labor exploitation ($\mu = 0.326$) across routes suggests \emph{systemic} patterns in migration risk. The route-specificity of waiting ($\mu = 0.125$) points to differences in transit infrastructure and border policy.

\section{Conclusion}

We introduced experiential intertextuality detection in migration narratives, a novel NLP task at the intersection of computational social science and discourse analysis. Our annotation-free pipeline, validated against expert judgments, shows that: (1)~experiential parallels across migration routes are real but hard to detect, with no single method exceeding $r = 0.38$ (56.6\% of the noise ceiling); (2)~a hybrid combining 31~features reaches $r = 0.45$ (68.6\% of ceiling); (3)~departure experiences are the most universally shared; (4)~LLM scores subsume embeddings; and (5)~prompting effects are model-dependent. Future work will explore fine-tuned models, richer discourse features, and experiential graphs linking shared experiences across the full corpus.


\section*{Limitations}

The corpus is small (108 narratives) with a strong route imbalance (99 Trans-Saharan vs.\ 9 Balkan), which may bias cross-route comparisons. The inter-annotator agreement ($\alpha = 0.273$), while consistent with task subjectivity, limits gold standard reliability; the noise ceiling analysis (\S\ref{sec:data}) shows that 31.4\% of the hybrid's gap to perfection is attributable to irreducible annotator noise. We evaluated only 7B-class LLMs at 4-bit quantization; larger models (70B+) or full-precision inference may yield different conclusions, and the effect of quantization on score distributions was not isolated. We did not fine-tune models because the annotation-free framing precludes training on expert labels by design; fine-tuned approaches are an important future direction. Our lexical baselines do not apply French lemmatization or morphological normalization, which may understate their potential given French's rich inflectional morphology. The 533 single-annotated pairs may carry annotator-specific bias; however, the correlation of automated methods on dual-annotated vs.\ single-annotated subsets shows consistent patterns, and both annotators' score distributions are similar (means: 0.389 vs.\ 0.405). The POS features are limited to shallow patterns; richer event representations (semantic role labels, predicate-argument structures, or frame-semantic features) could better capture experiential content and deserve exploration in future work. The context-aware features exploit only positional information; discourse-level features (coreference, causal chains) remain unexplored. For the hybrid model, we mitigate data leakage via narrative-level GroupKFold (\S\ref{sec:methods}); TF-IDF weights are fitted on the full 108-narrative corpus (not on CV folds), analogous to using a pretrained model, and all embedding/LLM scores are computed without access to expert labels. Finally, the theme lexicon is used both as a feature and for pair labeling; while the two serve different purposes (8 word-set scoring vs.\ 15-category labeling) and labeling was not used to select annotation pairs, we acknowledge this dual role. Empirically, the number of theme labels per pair is \emph{negatively} correlated with expert scores ($r = -0.148$, $p < 0.001$), ruling out the concern that thematic stratification biases toward high-scoring pairs. The partial correlation analysis (\S\ref{sec:results}) further confirms that LLM and lexicon signals are not reducible to theme-based distributional effects.

\section*{Ethical Considerations}
\label{sec:ethics}

The narratives were collected from minors and young adults at transit points during their migratory journeys, a context of extreme vulnerability. Interviews were conducted by trained researchers from transit countries and local associations, with a relationship of trust established through time and the use of the language spoken by the minors. All life stories are fully anonymized: names are replaced with codes, and identifying details removed. \textbf{The dataset will not be publicly released} due to the sensitive and personal nature of the content. Access may be granted to qualified researchers upon ethical review, in coordination with the ANR HYCI project partners. We release the evaluation pipeline code to support methodological reproducibility.

\section*{Acknowledgements}
This work has benefited from the support of the Hauts-de-France region, ANR HYCI Project (ANR-22-CE55-0010) of the French National Research Agency, CRIL-Lab CNRS, and Artois University.

\bibliography{references}

\begin{thebibliography}{24}
\providecommand{\natexlab}[1]{#1}

\bibitem[{Agirre et~al.(2016)Agirre, Banea, Cer, Diab, Gonzalez-Agirre,
  Mihalcea, Rigau, and Wiebe}]{agirre2016semeval}
Eneko Agirre, Carmen Banea, Daniel Cer, Mona Diab, Aitor Gonzalez-Agirre, Rada
  Mihalcea, German Rigau, and Janyce Wiebe. 2016.
\newblock {S}em{E}val-2016 task 1: Semantic textual similarity, monolingual and
  cross-lingual evaluation.
\newblock In \emph{Proceedings of SemEval}, pages 497--511. ACL.

\bibitem[{Bacon(2022)}]{bacon2022fabrique}
Lucie Bacon. 2022.
\newblock \emph{La fabrique du parcours migratoire sur la route des {B}alkans:
  Co-construction des r{\'e}cits et {\'e}critures (carto)graphiques}.
\newblock Ph.D. thesis, Universit{\'e} de Poitiers.

\bibitem[{Bamman et~al.(2013)Bamman, O'Connor, and Smith}]{bamman2013learning}
David Bamman, Brendan O'Connor, and Noah~A Smith. 2013.
\newblock Learning latent personas of film characters.
\newblock In \emph{Proceedings of ACL}, pages 352--361. ACL.

\bibitem[{Caselli and Vossen(2017)}]{caselli2017event}
Tommaso Caselli and Piek Vossen. 2017.
\newblock The event {S}tory{L}ine corpus: A new benchmark for causal and
  temporal relation extraction.
\newblock In \emph{Proceedings of the Events and Stories in the News Workshop},
  pages 77--86. ACL.

\bibitem[{Cer et~al.(2017)Cer, Diab, Agirre, Lopez-Gazpio, and
  Specia}]{cer2017semeval}
Daniel Cer, Mona Diab, Eneko Agirre, I{\~n}igo Lopez-Gazpio, and Lucia Specia.
  2017.
\newblock {S}em{E}val-2017 task 1: Semantic textual similarity multilingual and
  crosslingual focused evaluation.
\newblock In \emph{Proceedings of SemEval}, pages 1--14. ACL.

\bibitem[{Chambers and Jurafsky(2008)}]{chambers2008unsupervised}
Nathanael Chambers and Dan Jurafsky. 2008.
\newblock Unsupervised learning of narrative event chains.
\newblock In \emph{Proceedings of ACL}, pages 789--797. ACL.

\bibitem[{Conneau et~al.(2020)Conneau, Khandelwal, Goyal, Chaudhary, Wenzek,
  Guzm{\'a}n, Grave, Ott, Zettlemoyer, and Stoyanov}]{conneau2020unsupervised}
Alexis Conneau, Kartikay Khandelwal, Naman Goyal, Vishrav Chaudhary, Guillaume
  Wenzek, Francisco Guzm{\'a}n, Edouard Grave, Myle Ott, Luke Zettlemoyer, and
  Veselin Stoyanov. 2020.
\newblock Unsupervised cross-lingual representation learning at scale.
\newblock In \emph{Proceedings of ACL}, pages 8440--8451. ACL.

\bibitem[{Cybulska and Vossen(2014)}]{cybulska2014using}
Agata Cybulska and Piek Vossen. 2014.
\newblock Using a sledgehammer to crack a nut? lexical diversity and event
  coreference resolution.
\newblock In \emph{Proceedings of LREC}, pages 4545--4552.

\bibitem[{Forstall et~al.(2015)Forstall, Scheirer, Bamman, and
  Crane}]{forstall2015modeling}
Christopher Forstall, Walter Scheirer, David Bamman, and Gregory Crane. 2015.
\newblock Modeling the scholars: Detecting intertextuality through enhanced
  word-level n-gram matching.
\newblock \emph{Digital Scholarship in the Humanities}, 30(4):503--515.

\bibitem[{Gilardi et~al.(2023)Gilardi, Alizadeh, and
  Kubli}]{gilardi2023chatgpt}
Fabrizio Gilardi, Meysam Alizadeh, and Ma{\"e}l Kubli. 2023.
\newblock {ChatGPT} outperforms crowd workers for text-annotation tasks.
\newblock \emph{Proceedings of the National Academy of Sciences},
  120(30):e2305016120.

\bibitem[{Hussain et~al.(2018)Hussain, Bandeli, Al-khateeb, and
  Agarwal}]{hussain2018analyzing}
Muhammad~Nihal Hussain, Kevin~K Bandeli, Samer Al-khateeb, and Nitin Agarwal.
  2018.
\newblock Analyzing shift in narratives regarding migrants in {E}urope via
  blogosphere.
\newblock In \emph{Proceedings of the International Conference on Social
  Computing, Behavioral-Cultural Modeling and Prediction}, pages 181--190.
  Springer.

\bibitem[{Ing et~al.(2025)Ing, Delorme, Jabbour, Robin, and
  Sais}]{ing2025textmining}
David Ing, Fabien Delorme, Said Jabbour, Nelly Robin, and Lakhdar Sais. 2025.
\newblock Text mining from migration narratives.
\newblock In \emph{Proceedings of ECML-PKDD}.

\bibitem[{Jiang et~al.(2023)Jiang, Sablayrolles, Mensch, Bamford, Chaplot,
  de~las Casas, Bressand, Lengyel, Lample, Saulnier et~al.}]{jiang2023mistral}
Albert~Q Jiang, Alexandre Sablayrolles, Arthur Mensch, Chris Bamford,
  Devendra~Singh Chaplot, Diego de~las Casas, Florian Bressand, Gianna Lengyel,
  Guillaume Lample, Lucile Saulnier, and 1 others. 2023.
\newblock Mistral 7{B}.
\newblock \emph{arXiv preprint arXiv:2310.06825}.

\bibitem[{Kristeva(1969)}]{kristeva1969semeiotike}
Julia Kristeva. 1969.
\newblock \emph{S{\'e}m{\'e}iotik{\`e}: Recherches pour une s{\'e}manalyse}.
\newblock Seuil, Paris.

\bibitem[{Mann and Thompson(1988)}]{mann1988rhetorical}
William~C Mann and Sandra~A Thompson. 1988.
\newblock Rhetorical structure theory: Toward a functional theory of text
  organization.
\newblock \emph{Text}, 8(3):243--281.

\bibitem[{Martin et~al.(2020)Martin, Muller, Su{\'a}rez, Dupont, Romary, de~la
  Clergerie, Seddah, and Sagot}]{martin2020camembert}
Louis Martin, Benjamin Muller, Pedro Javier~Ortiz Su{\'a}rez, Yoann Dupont,
  Laurent Romary, {\'E}ric~Villemonte de~la Clergerie, Djam{\'e} Seddah, and
  Beno{\^i}t Sagot. 2020.
\newblock {C}amem{BERT}: a tasty {F}rench language model.
\newblock In \emph{Proceedings of ACL}, pages 7203--7219. ACL.

\bibitem[{Mostafazadeh et~al.(2016)Mostafazadeh, Chambers, He, Parikh, Batra,
  Vanderwende, Kohli, and Allen}]{mostafazadeh2016corpus}
Nasrin Mostafazadeh, Nathanael Chambers, Xiaodong He, Devi Parikh, Dhruv Batra,
  Lucy Vanderwende, Pushmeet Kohli, and James Allen. 2016.
\newblock A corpus and cloze evaluation for deeper understanding of commonsense
  stories.
\newblock In \emph{Proceedings of NAACL-HLT}, pages 839--849. ACL.

\bibitem[{{\"O}zt{\"u}rk and Ayvaz(2018)}]{ozturk2018sentiment}
Nihan {\"O}zt{\"u}rk and Serkan Ayvaz. 2018.
\newblock Sentiment analysis on {T}witter: A text mining approach to the
  {S}yrian refugee crisis.
\newblock \emph{Telematics and Informatics}, 35(1):136--147.

\bibitem[{Plekhanov et~al.(2023)Plekhanov, Kassner, Popat, Martin, Merello,
  Kozlovskii, Dreyer, and Cancedda}]{plekhanov2023multilingual}
Mikhail Plekhanov, Nora Kassner, Kashyap Popat, Louis Martin, Simone Merello,
  Boris Kozlovskii, Fabio~A Dreyer, and Nicola Cancedda. 2023.
\newblock Multilingual end to end entity linking.
\newblock In \emph{Proceedings of ACL}, pages 3512--3527. ACL.

\bibitem[{Prasad et~al.(2008)Prasad, Dinesh, Lee, Miltsakaki, Robaldo, Joshi,
  and Webber}]{prasad2008penn}
Rashmi Prasad, Nikhil Dinesh, Alan Lee, Eleni Miltsakaki, Livio Robaldo,
  Aravind Joshi, and Bonnie Webber. 2008.
\newblock The {P}enn discourse {T}ree{B}ank 2.0.
\newblock In \emph{Proceedings of LREC}, pages 2961--2968.

\bibitem[{{Qwen Team}(2024)}]{qwen2024}
{Qwen Team}. 2024.
\newblock Qwen2.5 technical report.
\newblock \emph{arXiv preprint arXiv:2412.15115}.

\bibitem[{Reagan et~al.(2016)Reagan, Mitchell, Kiley, Danforth, and
  Dodds}]{reagan2016emotional}
Andrew~J Reagan, Lewis Mitchell, Dilan Kiley, Christopher~M Danforth, and
  Peter~Sheridan Dodds. 2016.
\newblock The emotional arcs of stories are dominated by six basic shapes.
\newblock \emph{EPJ Data Science}, 5(1):1--12.

\bibitem[{Reimers and Gurevych(2019)}]{reimers2019sentencebert}
Nils Reimers and Iryna Gurevych. 2019.
\newblock Sentence-{BERT}: Sentence embeddings using siamese {BERT}-networks.
\newblock In \emph{Proceedings of EMNLP-IJCNLP}, pages 3982--3992. ACL.

\bibitem[{Robin(2014)}]{robin2014migrations}
Nelly Robin. 2014.
\newblock Migrations, observatoire et droit: Complexit{\'e} du syst{\`e}me
  migratoire ouest-africain.
\newblock In \emph{HdR}. Universit{\'e} de Poitiers.

\end{thebibliography}

\appendix

\section{Prompt Templates}
\label{app:prompts}

\paragraph{Zero-shot system prompt (French):}
\begin{quote}\small
\textit{Tu es un expert en \'etudes migratoires et analyse de discours. \'Evalue le degr\'e d'intertextualit\'e exp\'erientielle entre deux phrases issues de r\'ecits de migration fran\c{c}ais provenant de routes migratoires diff\'erentes. L'intertextualit\'e exp\'erientielle = exp\'eriences v\'ecues partag\'ees, \'echos th\'ematiques, situations parall\`eles. \'Echelle: 0.0 (aucun lien) \`a 1.0 (exp\'eriences quasi-identiques). R\'eponds UNIQUEMENT avec un JSON: \{"score": <float>\}}
\end{quote}

\paragraph{CoT system prompt (French):}
\begin{quote}\small
\textit{[\ldots] Raisonne \'etape par \'etape: 1.\ Identifie le th\`eme de chaque phrase. 2.\ Compare: s'agit-il d'exp\'eriences parall\`eles? 3.\ \'Evalue la force du lien. R\'eponds avec un JSON: \{``theme1'': ``\ldots'', ``theme2'': ``\ldots'', ``shared\_experience'': ``\ldots'', ``score'': <float>\}}
\end{quote}

\paragraph{Few-shot examples (3 demonstrations):}
Selected from expert-annotated pairs to cover low ($\sim$0.1), medium ($\sim$0.4), and high ($\sim$0.8) scores, drawn from pairs \emph{not} in the evaluation set. Each example shows the two sentences and the expert consensus score.

\section{Annotation Sampling Protocol}
\label{app:sampling}

The 816 expert-annotated pairs were sampled from the full auto-generated pool with stratification by thematic category to ensure coverage across all 15~themes (annotation was conducted via a multi-user Streamlit application). In practice, 815 of 816 annotated pairs are cross-route (one Balkan, one Trans-Saharan), reflecting the primary research question of cross-corridor experiential echoes. The single intra-route pair entered through an edge case in the sampling procedure. The thematic distribution of annotated pairs is: \emph{attente\_stagnation} (123), \emph{autre} (115), \emph{passeur\_exploitation} (93), \emph{violence\_police} (82), \emph{travers\'ee\_dangereuse} (79), \emph{exploitation\_travail} (69), \emph{solidarit\'e} (66), \emph{document\_fraude} (57), \emph{famille\_s\'eparation} (52), \emph{r\^eve\_football} (49), \emph{mineur\_seul} (46), \emph{discrimination\_racisme} (43), \emph{d\'etention\_camp} (42), \emph{motivation\_d\'epart} (39), \emph{mort\_danger\_vital} (30).

\end{document}